\documentclass{article}
\usepackage{rise_conference,times}

\usepackage{amsmath,amsfonts,bm}

\def\eqref#1{equation~\ref{#1}}

\def\1{\bm{1}}

\DeclareMathAlphabet{\mathsfit}{\encodingdefault}{\sfdefault}{m}{sl}
\SetMathAlphabet{\mathsfit}{bold}{\encodingdefault}{\sfdefault}{bx}{n}

\usepackage{graphicx}
\usepackage{flafter}
\usepackage{booktabs}
\usepackage{multirow}
\usepackage{capt-of}
\usepackage[table]{xcolor}
\usepackage{hyperref}
\usepackage{url}
\usepackage{fontawesome5}
\usepackage{placeins}

\definecolor{citationblue}{RGB}{31,119,180}
\definecolor{projectpink}{RGB}{255,0,128}
\hypersetup{
    colorlinks=true,
    linkcolor=citationblue,
    citecolor=citationblue,
    urlcolor=citationblue
}

\definecolor{rankfirst}{RGB}{255,203,153}
\definecolor{ranksecond}{RGB}{255,243,205}
\definecolor{rankthird}{RGB}{220,235,247}
\newcommand{\bestresult}[1]{\cellcolor{rankfirst}\textbf{#1}}
\newcommand{\secondresult}[1]{\cellcolor{ranksecond}#1}

\newcommand{\tableonebest}[1]{\cellcolor{rankfirst}\textbf{#1}}
\newcommand{\tableonesecond}[1]{\cellcolor{ranksecond}#1}

\title{Learning Structural Illumination for\\ Unsupervised Low-Light Enhancement}

\author{Tianle Du$^{1}$,
Peiyuan He$^{1}$,
Hainuo Wang$^{1}$,
Tianxiu Yu$^{2}$,
Xiaojie Guo$^{1,}$\thanks{Corresponding author.}\\
$^{1}$Tianjin University \ 
$^{2}$Dunhuang Academy \\
\{dutianle, peiyuan\_he, hainuo\}@tju.edu.cn, yutx@dua.ac.cn, xj.max.guo@gmail.com \\
}

\iclrfinalcopy
\begin{document}
\maketitle

\begin{abstract}
Existing unsupervised low-light image enhancement (LLIE) methods often estimate illumination directly from the entire low-light input, without separating its spatially varying illumination pattern, termed relative illumination structure, from the absolute exposure level or preventing unreliable low signal-to-noise ratio regions from biasing the estimate. Moreover, fixed exposure targets impose a scene-agnostic enhancement criterion, limiting adaptation across diverse lighting conditions. Inspired by the spatial propagation of light, we propose a \textit{Relative Illumination Structure Estimation} (RISE) framework that decouples relative illumination structure from absolute exposure and infers it from reliable bright regions, enabling interpretable and robust enhancement. For scene-adaptive exposure adjustment, we further propose a \textit{Dual-Metering Exposure Reference} derived from each input, allowing RISE to adapt the enhancement strength to individual scenes and generalize across diverse lighting conditions. Extensive benchmark and real-world generalization experiments show that RISE achieves state-of-the-art performance among unsupervised LLIE methods while producing visually natural results.

\textbf{Project page:} \href{https://dutianle.ltd/RISE}{\textcolor{projectpink}{https://dutianle.ltd/RISE}}\,\href{https://dutianle.ltd/RISE}{\textcolor{black}{\faGithub}}
\end{abstract}

\begin{center}
    \centering
    \includegraphics[width=\textwidth]{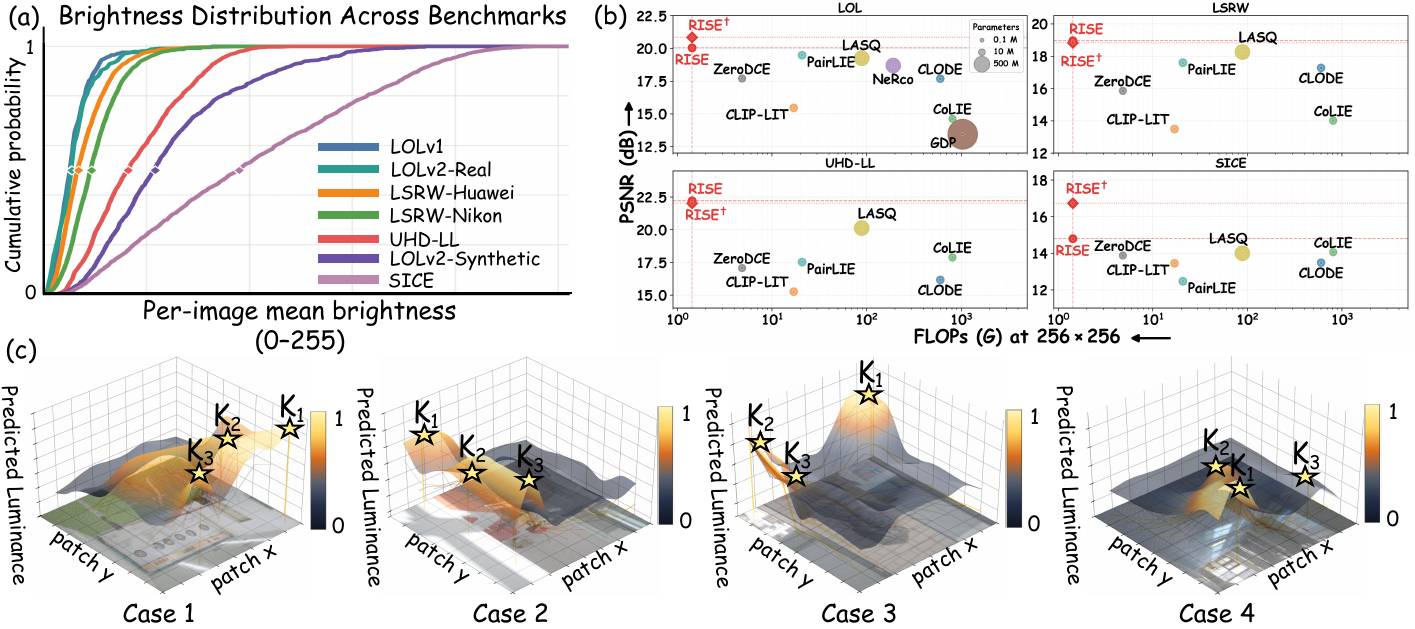}
    \captionof{figure}{(a) Empirical cumulative distribution functions (CDFs) of per-image mean brightness across seven benchmarks. (b) RISE consistently delivers a superior PSNR--FLOPs trade-off across all benchmarks, achieving higher PSNR than prior SOTA methods with the fewest FLOPs. (c) Joint visualization of relative illumination structure and absolute exposure: surface shape encodes relative illumination, elevation encodes absolute exposure, and stars mark selected KICs.}
    \label{fig:performance}
\end{center}

\section{Introduction}

Low-light image enhancement (LLIE) is a fundamental vision task that aims to recover well-exposed images from underexposed observations, benefiting both human perception and downstream understanding. However, supervised methods rely on costly, hard-to-scale, and capture-biased paired data, motivating unsupervised LLIE that learns from low-light images without paired ground truth.

In recent years, deep learning has substantially advanced unsupervised LLIE.
However, without paired ground-truth references, there is no direct supervisory signal for target exposure. To compensate, many unsupervised methods~\citep{guo2020zero,li2025interpretable}
are trained with fixed exposure values obtained from statistical priors as surrogate supervision. Under such objectives, scene-agnostic supervision tends to bias enhanced results toward a uniform brightness level, overlooking variations among different scenes and preventing image-adaptive enhancement. Consequently, such methods fail to generalize well across the diverse brightness conditions illustrated in Fig.~\ref{fig:performance}(a).

\begin{figure}[t]
    \centering
    \includegraphics[width=\textwidth]{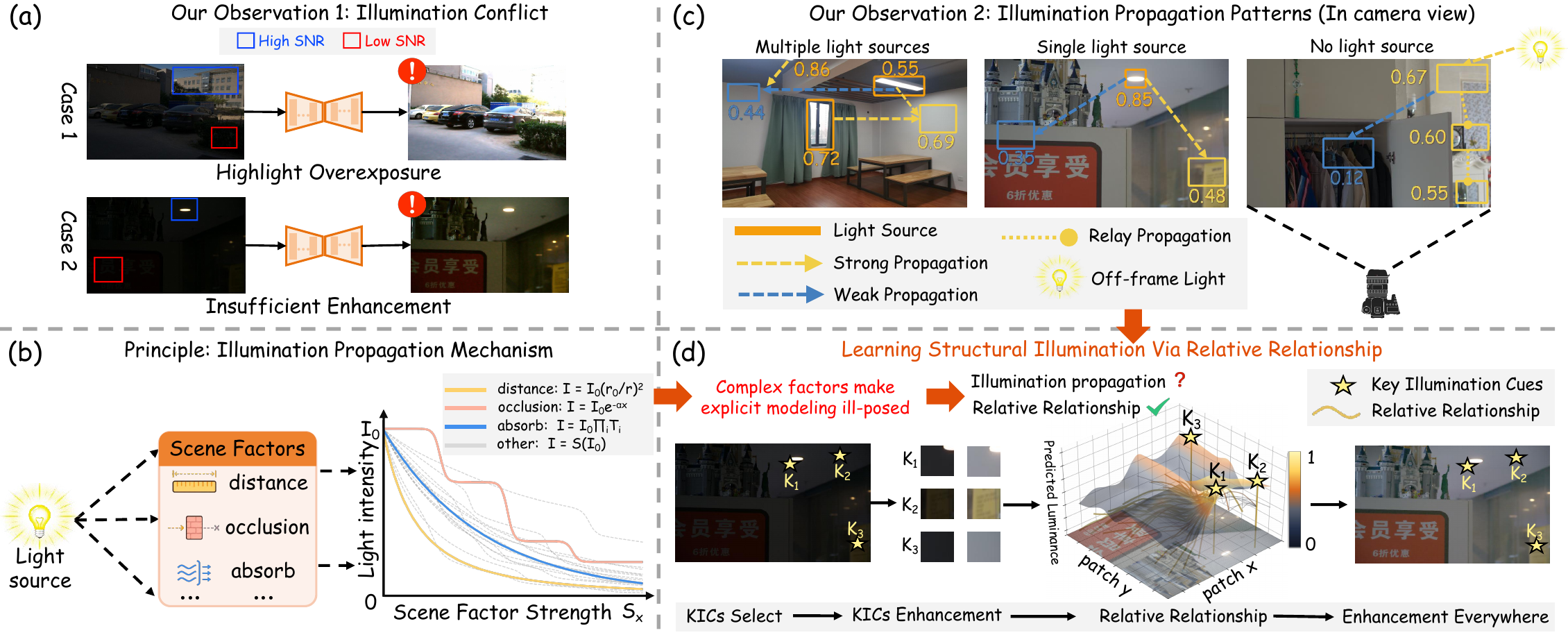}
    \caption{Motivation of RISE. (a) Existing methods show unstable enhancement. (b) Light intensity from a source
    attenuates with scene factors. (c) Relative illumination
    propagation patterns in normal-light images~(numbers denote regional brightness). (d) Our insight: RISE learns illumination structure from relative relationships to key illumination cues.}
    \label{fig:motivation}
\end{figure}

Another key challenge is the interference of low signal-to-noise ratio (SNR) regions in illumination estimation. Under severe under-exposure, pixel observations are dominated by sensor noise and contain only weak scene signals. Nevertheless, most methods~\citep{guo2016lime,zhang2019kindling,jiang2024lightendiffusion} estimate illumination by treating the whole low-light image as uniformly reliable, leading to unstable enhancement. As shown in Fig.~\ref{fig:motivation}(a), the model may fail to brighten extremely dark regions while preventing highlight overexposure. 
Although recent supervised method SNR-Aware~\citep{xu2022snr} exploits long-range dependencies between low-SNR and high-SNR regions to alleviate this issue, it still treats unreliable regions as useful features, leaving their interference unresolved.

Beyond these challenges, the representation of scene illumination itself is also critical. \textit{Rather than a single quantity, illumination comprises a spatially varying structure that captures relative brightness relationships and a global scale that controls the absolute exposure level}~\citep{zhan2021emlight,garon2019fast}. Ideally, LLIE should therefore be formulated as $\mathbf{S}=g_{\theta}(\mathbf{I}_{\mathrm{low}})$, $\eta=h_{\phi}(\mathbf{I}_{\mathrm{low}})$, and $\hat{\mathbf{I}}=\mathcal{E}(\mathbf{I}_{\mathrm{low}};\mathbf{S},\eta)$, where $\mathbf{S}$ captures the relative illumination structure and $\eta$ independently determines the desired exposure level. Existing methods, however, typically predict a single enhancement representation $\mathbf{Z}=f_{\theta}(\mathbf{I}_{\mathrm{low}})$ and produce the enhanced result as $\hat{\mathbf{I}}=\mathcal{E}(\mathbf{I}_{\mathrm{low}};\mathbf{Z})$, where $\mathbf{Z}$ may be an illumination map~\citep{xie2024residual}, an enhancement curve~\citep{guo2020zero}, or the enhanced image itself~\citep{wang2024zero}. Such a formulation entangles spatial illumination structure with absolute exposure, leaving their distinct roles neither explicitly identifiable nor independently controllable.


These limitations motivate us to seek illumination cues that are both structured and reliable. Such cues naturally arise from how light is distributed in real scenes. As shown in Fig.~\ref{fig:motivation}(b), light propagates from its source with spatially varying strength due to distance, occlusion, absorption, and other scene factors. Similar patterns can be observed in normal-light images, where brightness variations exhibit relative propagation from well-lit regions to darker areas under diverse lighting conditions, as shown in Fig.~\ref{fig:motivation}(c). However, directly modeling the full propagation mechanism from a single low-light image is ill-posed because the underlying scene factors are complex and unobserved. But this view suggests that illumination can be inferred through relative relationships to reliable bright regions, with the relative relationships themselves reflecting the underlying illumination structure, as illustrated in Fig.~\ref{fig:motivation}(d). These regions lie on strong light-propagation paths, and we refer to them as Key Illumination Cues~(KICs). Based on this insight, we propose a Relative Illumination Structure Estimation (RISE) framework that learns illumination structure by modeling relative relationships to KICs and decouples it from the absolute exposure level, reducing unstable enhancement.

Meanwhile, to achieve scene-adaptive exposure adjustment, we further propose a Dual-Metering Exposure Reference (DMER) derived from each input. DMER replaces fixed exposure values and training-data statistics with an input-dependent reference that accounts for both global exposure demand and relative illumination structure. Guided by this reference, RISE adjusts exposure for individual images rather than memorizing dataset-specific brightness preferences, improving generalization across datasets and real-world scenes.

Our main contributions are summarized as follows:
\begin{itemize}
    \item We recast unsupervised LLIE from a structured and relational illumination perspective, providing a new problem formulation for low-light image enhancement.
    \item We propose a Relative Illumination Structure Estimation (RISE) framework that decouples the relative illumination structure from absolute exposure and infers illumination from sparse yet reliable bright regions for interpretable and robust enhancement. 
    \item We design Dual-Metering Exposure Reference (DMER), an input-dependent exposure reference that captures global and relative illumination cues, improving cross-dataset and real-world generalization.
\end{itemize}

Comprehensive experiments on multiple benchmarks show that RISE achieves
state-of-the-art performance, as summarized in Fig.~\ref{fig:performance}(b), and more natural enhancement, while extensive
ablation studies verify the rationality and effectiveness of the proposed design.

\section{Related Work}


\noindent\textbf{Unsupervised LLIE.} Supervised LLIE methods~\citep{guo2023low,du2026anchor} learn from paired low-/normal-light images, but their dependence on costly paired data limits scalability and motivates unsupervised enhancement. Existing unsupervised methods seek alternative supervision signals to replace paired references. Li et al.~\citep{li2025interpretable} control exposure by enforcing image patches toward a fixed brightness value, ResQ-Net~\citep{xie2024residual} uses the gray-world assumption to drive exposure adjustment, LightenDiffusion~\citep{jiang2024lightendiffusion} and related diffusion-based methods learn latent normal-light priors from unpaired data,  CLIP-LIT~\citep{liang2023iterative} trains the enhancement network by maximizing similarity to positive language prompts, and PairLIE~\citep{fu2023learning} mines supervision from paired low-light images of the same scene under different exposures. Although these methods have achieved promising results, their supervision still depends on fixed exposure targets, training-data statistics, or external priors. As a result, they tend to learn dataset-dependent enhancement preferences rather than scene-adaptive exposure adjustment, leading to limited generalization under diverse lighting conditions.

\begin{figure}[t]
    \centering
    \includegraphics[width=\textwidth]{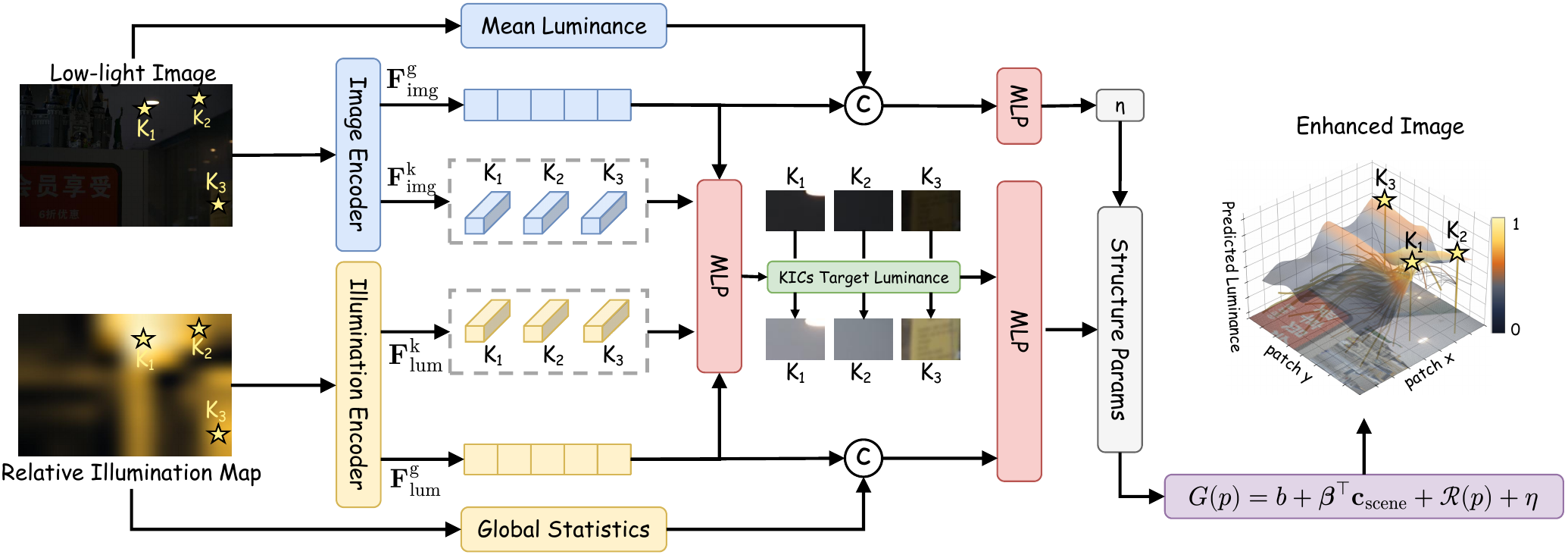}
    \caption{Overview of RISE. RISE jointly encodes image appearance and relative illumination, using spatially dispersed KICs as reliable anchors to estimate illumination structure while independently predicting absolute exposure for scene-adaptive enhancement, enabling reliable structural modeling and adaptive exposure control across diverse scenes.}
    \label{fig:rise}
\end{figure}

\noindent\textbf{Illumination Modeling for LLIE.} In LLIE, the illumination is explicitly or implicitly modeled. Retinex-based methods model illumination with dense maps. LIME~\citep{guo2016lime} constructs an illumination map from channel-wise maxima and refines it with structure-aware smoothing. RUAS~\citep{liu2021retinex} searches for an illumination estimation network from a compact architecture space through architecture search. IRLE~\citep{he2026internally} guides illumination estimation with an inverse gain map from the input image. Another line of work models enhancement as iterative lightening curves. Zero-DCE~\citep{guo2020zero} predicts pixel-wise parameters of high-order curves, while CLODE~\citep{jung2025continuous} formulates curve adjustment as a continuous dynamic system and solves it with neural ordinary differential equations. Implicit representations have also been explored for enhancement. LLNeRF~\citep{wang2023lighting} enhances brightness in 3D neural radiance fields, NeRCo~\citep{yang2023implicit} uses implicit neural representations to normalize low-light inputs under different degradation levels, and NoiSER~\citep{zhang2024noise} implicitly learns brightness enhancement from zero-mean Gaussian noise. Although these methods represent illumination in different forms, they do not explicitly treat scene illumination as a structured quantity decoupled from absolute exposure, nor do they account for the adverse influence of unreliable low-SNR regions on illumination representation.

\section{Method}

\begin{figure}[t]
    \centering
    \includegraphics[width=0.95\linewidth]{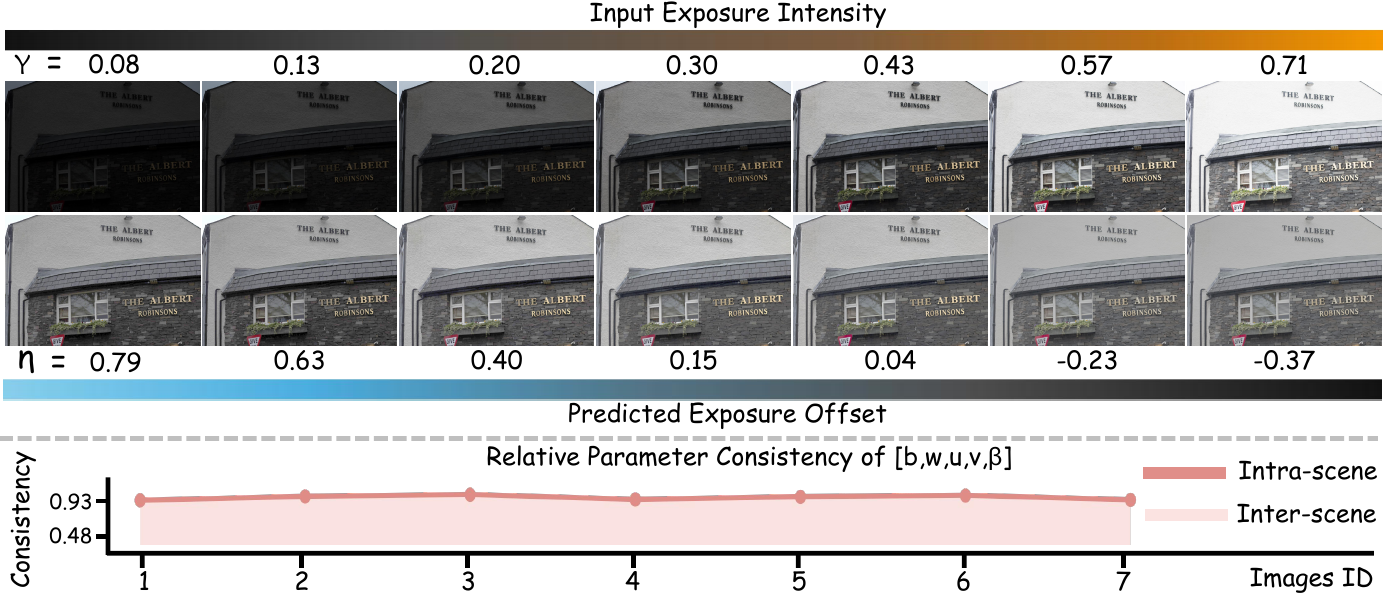}
    \caption{Validation of illumination structure and exposure decoupling on SICE. Under different exposures of the same scene, RISE maintains stable relative parameters while adjusting the absolute exposure, which demonstrates its ability to both enhance underexposed inputs and suppress overexposed scenes. In contrast, relative parameter consistency across different scenes remains low.}
    \label{fig:scene}
\end{figure}

As schematically depicted in Fig.~\ref{fig:rise}, RISE learns scene illumination structure by modeling relative relationships to KICs while separately controlling the absolute exposure level. Given a low-light input $\mathbf{I}_{\mathrm{low}}$, RISE first derives a relative illumination map $\mathbf{Z}$, where global brightness is removed. It then selects $N$ spatially sparse bright regions as KICs and first estimates their enhanced luminance, as their higher SNR enables more reliable luminance prediction. Based on these KICs, a relation modeling branch estimates their relative relationships to other regions, forming the relative illumination structure $\mathbf{S}$ that relates KICs to the rest of the scene. In parallel, an exposure control branch predicts an image-level exposure offset $\eta$, which determines the overall output brightness. The whole framework is trained with unsupervised objectives guided by DMER.

\subsection{Relative Illumination Structure Estimation}
Since the light source and propagation factors are not directly observable, RISE uses KICs as observable anchors for estimating the relative illumination structure. We first convert the input $\mathbf{I}_{\mathrm{low}}$ to the luminance channel
$\mathbf{Y}=0.299\,\mathbf{I}_{\mathrm{low}}^{\mathrm{R}}
 +0.587\,\mathbf{I}_{\mathrm{low}}^{\mathrm{G}}
 +0.114\,\mathbf{I}_{\mathrm{low}}^{\mathrm{B}}$. Then, $\mathbf{Y}$ is
average-pooled with an $h \times w$ window to obtain patch luminance
$\mathbf{P}$, where each element corresponds to the mean luminance of a local
patch. To remove global brightness and retain relative illumination, we define the relative illumination map
$\mathbf{Z}$:
\begin{equation}
    \mathbf{Z} = \log\!\left(
    \frac{\mathbf{P}}{\operatorname{mean}(\mathbf{P})+\epsilon}
    \right).
    \label{eq:zmap}
\end{equation}
Here, $\epsilon$ ensures numerical stability, while $\log(\cdot)$ maps normalized brightness ratios to signed offsets relative to the unit-ratio reference and compresses their dynamic range.

We then select $N$ spatially dispersed KICs from the brightest patches in $\mathbf{Z}$, denoted by $\{k_n\}_{n=1}^{N}$. An MLP predicts their target luminance values $\hat{\boldsymbol{\ell}}$, which serve as reliable anchors for relational illumination modeling. RISE builds a structured field by positioning each patch relative to the KICs in illumination and spatial domains. The spatial relation is motivated by the attenuation of light along its propagation path. However, the true propagation distance is three-dimensional and cannot be reliably recovered from a single image. We therefore use image-plane distance only as a soft attenuation surrogate rather than a physical measurement. For patch $p$ and KIC $k_n$, we define:
\begin{equation}
    \delta_n(p)=Z_{k_n}-Z_p,\qquad
    \rho_n(p) = \frac{1}{1+d(p,k_n)/D},
    \label{eq:kernel}
\end{equation}
where $d(p,k_n)$ is the patch-grid distance and $D$ is the grid diagonal.
The kernel $\rho_n(p)$ equals one at the KIC and decays monotonically with distance, providing a bounded attenuation prior rather than an inverse-square constraint on
image-plane geometry.  The discrepancy $\delta_n(p)$ retains the observed relative illumination change and implicitly reflects occlusion, absorption, and other effects beyond image-plane distance. Together, $\rho_n(p)$ and $\delta_n(p)$ define the relational field:
\begin{equation}
    \mathcal{R}(p)=\sum_{n=1}^{N}\big[
        w_n\,\delta_n(p)+u_n\,\rho_n(p)\,\delta_n(p)
        +v_n\,\rho_n(p)\,Z_{k_n}\big].
    \label{eq:relation}
\end{equation}
The unmodulated discrepancy $\delta_n(p)$ preserves the observed relation when image-plane
distance is unreliable. Its distance-weighted form $\rho_n(p)\delta_n(p)$ combines
the attenuation prior with the implicit scene factors encoded in $\delta_n(p)$, while
$\rho_n(p)Z_{k_n}$ transfers the illumination strength of the KIC under the
same decay. The image-adaptive
coefficients $w_n$, $u_n$, and $v_n$ balance these cues for each KIC, allowing
multiple well-lit regions to jointly shape the relational field at each patch.

Although $\mathcal{R}(p)$ captures pairwise KIC-to-patch relations, similar local
relations may arise from different scene-level illumination patterns. We therefore
introduce $\mathbf{c}_{\mathrm{scene}}$ to encode these differences by summarizing global statistics of $\mathbf{Z}$. The affine projection
$\boldsymbol{\beta}^{\top}\mathbf{c}_{\mathrm{scene}}+b$ broadcasts this scene
context to all regions. Together, the relational field and scene context define
the relative illumination structure, while $\eta$ independently controls the
absolute exposure. Since illumination correction changes luminance by rescaling rather than shifting it, we parameterize the correction in log-gain space, where separate gain factors compose additively:
\begin{equation}
    \hat{G}(p)=
    \underbrace{b+\boldsymbol{\beta}^{\top}\mathbf{c}_{\mathrm{scene}}
    +\mathcal{R}(p)}_{\text{relative structure }S(p)}
    +\underbrace{\eta}_{\text{absolute exposure}}.
    \label{eq:gain}
\end{equation}

The relative parameters
$\Theta=\big(b,\boldsymbol{\beta},\{w_n,u_n,v_n\}_{n=1}^{N}\big)$ and the
absolute exposure $\eta$ are predicted by separate MLPs. This decoupling
keeps structural parameters $\Theta$ stable across exposures of the same scene,
leaving $\eta$ to absorb exposure changes, as shown in Fig.~\ref{fig:scene}. The enhanced result is given by
\begin{equation}
    \hat{\mathbf{I}}=
    \mathbf{I}_{\mathrm{low}}\odot
    \exp\!\left(\operatorname{Up}(\hat{\mathbf{G}})\right).
\end{equation}

\begin{figure}[t]
    \centering
    \includegraphics[width=\linewidth]{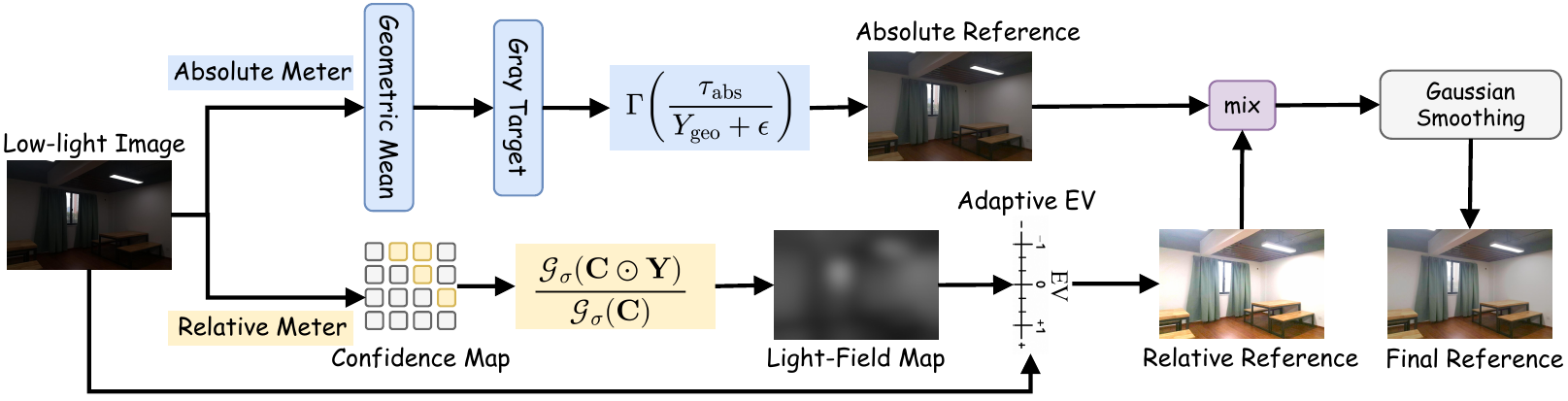}
    \caption{Overview of DMER. The absolute meter estimates global exposure, while the relative meter models spatial illumination through light-field propagation.}
    \label{fig:dmer}
\end{figure}

\subsection{Dual-Metering Exposure Reference}
Scene-adaptive enhancement should not rely on a fixed global target, since low-light images differ not only in exposure level but also in the spatial distribution of illumination. As shown in Fig.~\ref{fig:dmer}, DMER estimates the enhancement reference from each input with two complementary exposure meters. The absolute meter captures the global exposure trend through a conservative gray-world estimate, while the relative meter adjusts the reference according to spatial illumination differences. The absolute meter constructs its reference by rescaling the global luminance toward a gray level:
\begin{equation}
    g_{\mathrm{abs}}=\Gamma\!\left(
    \frac{\tau_{\mathrm{abs}}}{Y_{\mathrm{geo}}+\epsilon}
    \right),\qquad
    \mathbf{T}_{\mathrm{abs}}=
    g_{\mathrm{abs}}\mathbf{I}_{\mathrm{low}}.
    \label{eq:dmerabs}
\end{equation}
Here, $Y_{\mathrm{geo}}$ is the geometric mean luminance of $\mathbf{Y}$, and $\tau_{\mathrm{abs}}$ is a fixed gray target across datasets in the normalized intensity range.  If this gain would push bright regions beyond the valid intensity range, $\Gamma(\cdot)$ clips it to the maximum gain that avoids overexposure.

Although this absolute reference provides a stable global baseline, it is spatially uniform and cannot reflect illumination imbalance within the image. DMER therefore complements it with a relative meter that uses the estimated illumination structure to modulate the reference. Specifically, we first estimate a luminance based confidence map as
${\mathbf{C}=\mathbf{Y}/(\mathbf{Y}+\operatorname{mean}(\mathbf{Y})+\epsilon)}$,
which assigns higher confidence to reliably illuminated regions. We then diffuse reliable luminance cues to construct a structure-aware light field:
\begin{equation}
    \mathbf{L} = \frac{\mathcal{G}_{\sigma}(\mathbf{C}\odot\mathbf{Y})}
                       {\mathcal{G}_{\sigma}(\mathbf{C})},
    \label{eq:dmerfield}
\end{equation}
where $\mathcal{G}_{\sigma}$ is Gaussian smoothing. This operation approximates the smooth spatial variation of illumination from reliable lit regions, allowing $\mathbf{L}$ to capture relative lighting strength across regions.

\begin{table*}[t]
    \centering
    \fontsize{8}{9.5}\selectfont
    \setlength{\tabcolsep}{0.25pt}
    \renewcommand{\arraystretch}{1.0}
    \caption{Quantitative comparison on LOL. FLOPs is tested on a single $256\times256$ image. The best and second-best results are highlighted in \colorbox{rankfirst}{\textbf{orange}} and \colorbox{ranksecond}{yellow}, respectively, with the best results also shown in boldface. SL and UL denote supervised and unsupervised learning.}
    \label{tab:quantitative-comparison-LOL}
    \begin{tabular*}{\textwidth}{@{\extracolsep{\fill}}clccccccccccc@{}}
        \toprule
        \multirow{2}{*}{\textbf{Type}}
        & \multirow{2}{*}{\textbf{Method}}
        & \multicolumn{3}{c}{\textbf{LOL-v1}}
        & \multicolumn{3}{c}{\textbf{LOLv2-Real}}
        & \multicolumn{3}{c}{\textbf{LOLv2-Synthetic}}
        & \multirow{2}{*}{\shortstack{\textbf{Params}\\\textbf{(M)}}}
        & \multirow{2}{*}{\shortstack{\textbf{FLOPs}\\\textbf{(G)}}} \\
        \cmidrule(lr){3-5} \cmidrule(lr){6-8} \cmidrule(lr){9-11}
        & & PSNR$\uparrow$ & SSIM$\uparrow$ & LPIPS$\downarrow$
        & PSNR$\uparrow$ & SSIM$\uparrow$ & LPIPS$\downarrow$
        & PSNR$\uparrow$ & SSIM$\uparrow$ & LPIPS$\downarrow$ & & \\
        \midrule
        \multirow{4}{*}{\rotatebox{90}{\textbf{SL}}}
        & MIRNet~\citep{zamir2020learning} & 20.57 & 0.772 & 0.254 & 21.28 & 0.792 & 0.354 & 21.74 & 0.878 & 0.138 & 31.79 & 760.10 \\
        & CIDNet~\citep{yan2025hvi} & 23.97 & 0.849 & 0.104 & 23.90 & 0.866 & 0.122 & 25.27 & 0.935 & 0.048 & 1.88 & 7.57 \\
        & Multinex~\citep{brateanu2026multinex} & 23.19 & 0.843 & 0.129 & 23.04 & 0.860 & 0.178 & 25.04 & 0.930 & 0.068 & 0.04 & 2.50 \\
        & M.-Nano~\citep{brateanu2026multinex} & 19.42 & 0.742 & 0.276 & 19.66 & 0.784 & 0.266 & 21.05 & 0.882 & 0.143 & 0.001 & 0.04 \\
        \midrule
        \multirow{13}{*}{\rotatebox{90}{\textbf{UL}}}
        & Zero-DCE~\citep{guo2020zero} & 14.86 & 0.559 & 0.335 & 18.06 & 0.574 & 0.313 & 17.76 & 0.816 & \tableonebest{0.168} & 0.08 & 4.83 \\
        & RUAS~\citep{liu2021retinex} & 16.41 & 0.500 & 0.270 & 15.33 & 0.488 & 0.310 & 13.40 & 0.644 & 0.364 & 0.003 & 0.83 \\
        & SCI~\citep{ma2022toward} & 14.78 & 0.522 & 0.339 & 17.30 & 0.534 & 0.308 & 15.43 & 0.748 & 0.233 & 0.001 & 0.06 \\
        & PairLIE~\citep{fu2023learning} & 19.51 & 0.736 & 0.248 & 19.88 & 0.778 & \tableonesecond{0.234} & \tableonesecond{19.07} & 0.797 & 0.230 & 0.34 & 20.81 \\
        & CLIP-LIT~\citep{liang2023iterative} & 12.39 & 0.493 & 0.382 & 15.18 & 0.529 & 0.369 & 16.19 & 0.775 & 0.204 & 0.28 & 16.96 \\
        & GDP~\citep{fei2023generative} & 15.82 & 0.541 & 0.339 & 14.40 & 0.494 & 0.362 & 12.12 & 0.497 & 0.356 & 552.81 & 1037.26 \\
        & NeRCo~\citep{yang2023implicit} & 19.74 & 0.743 & 0.234 & 19.66 & 0.717 & 0.271 & 17.59 & 0.734 & 0.301 & 23.05 & 191.06 \\
        & CoLIE~\citep{chobola2024fast} & 13.76 & 0.481 & 0.356 & 15.08 & 0.501 & 0.322 & 14.30 & 0.654 & 0.251 & 0.13 & 806.25 \\
        & Li et al.~\citep{li2025interpretable} & 19.82 & 0.751 & 0.254 & 20.35 & \tableonesecond{0.795} & 0.266 & 17.82 & 0.802 & 0.293 & 0.35 & 19.00 \\
        & CLODE~\citep{jung2025continuous} & 19.60 & 0.718 & 0.287 & 17.87 & 0.681 & 0.335 & 17.21 & 0.783 & 0.313 & 0.22 & 600.96 \\
        & LASQ~\citep{kong2026luminance} & 17.22 & 0.739 & 0.240 & 20.45 & 0.778 & 0.239 & 18.37 & 0.791 & 0.240 & 24.08 & 88.67 \\
        \cmidrule(l){2-13}
        & \textbf{RISE} & \tableonesecond{20.21} & \tableonesecond{0.772} & \tableonesecond{0.219} & \tableonesecond{21.33} & 0.780 & \tableonebest{0.215} & 18.72 & \tableonebest{0.846} & \tableonesecond{0.169} & 0.35 & 1.43 \\
        & \textbf{RISE}$^\dagger$ & \tableonebest{21.03} & \tableonebest{0.776} & \tableonebest{0.215} & \tableonebest{22.34} & \tableonebest{0.801} & 0.243 & \tableonebest{19.33} & \tableonesecond{0.838} & 0.200 & 0.35 & 1.43 \\
        \bottomrule
\end{tabular*}
\end{table*}

\begin{figure}[t]
    \centering
    \includegraphics[width=\textwidth]{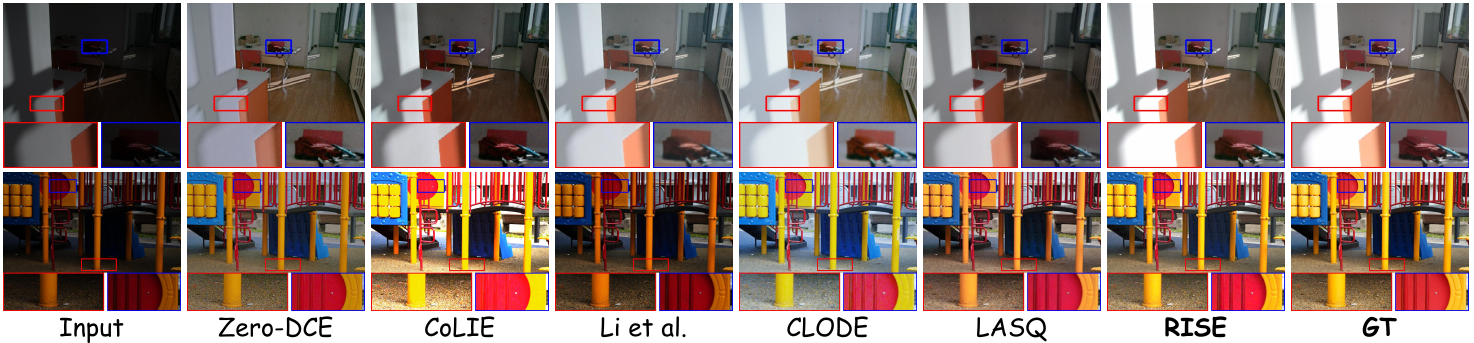}
    \caption{Visual comparisons of the enhanced results by different methods on LOL and UHD-LL.}
    \label{fig:lolresults}
    \vspace{-2mm}
\end{figure}

\begin{table*}[t]
    \centering
    \scriptsize
    \setlength{\tabcolsep}{1.1pt}
    \renewcommand{\arraystretch}{0.9}
    \caption{Quantitative comparison on LSRW, UHD-LL, and SICE.}
    \label{tab:quantitative-comparison-LSRW}
    \begin{tabular*}{\textwidth}{@{\extracolsep{\fill}}clcccccccccccc@{}}
        \toprule
        \multirow{2}{*}{\textbf{Type}}
        & \multirow{2}{*}{\textbf{Method}}
        & \multicolumn{3}{c}{\textbf{LSRW-Huawei}}
        & \multicolumn{3}{c}{\textbf{LSRW-Nikon}}
        & \multicolumn{3}{c}{\textbf{UHD-LL}}
        & \multicolumn{3}{c}{\textbf{SICE}} \\
        \cmidrule(lr){3-5} \cmidrule(lr){6-8} \cmidrule(lr){9-11} \cmidrule(lr){12-14}
        & & PSNR$\uparrow$ & SSIM$\uparrow$ & LPIPS$\downarrow$
        & PSNR$\uparrow$ & SSIM$\uparrow$ & LPIPS$\downarrow$
        & PSNR$\uparrow$ & SSIM$\uparrow$ & LPIPS$\downarrow$
        & PSNR$\uparrow$ & SSIM$\uparrow$ & LPIPS$\downarrow$ \\
        \midrule
        \multirow{4}{*}{\rotatebox{90}{\textbf{SL}}}
        & MIRNet~\citep{zamir2020learning} & 20.88 & 0.628 & 0.451 & 17.38 & 0.502 & 0.539 & 20.55 & 0.807 & 0.435 & - & - & - \\
        & CIDNet~\citep{yan2025hvi} & 20.71 & 0.618 & 0.359 & 17.14 & 0.503 & 0.300 & 23.80 & 0.875 & 0.209 & 18.60 & 0.611 & 0.407 \\
        & Multinex~\citep{brateanu2026multinex} & 21.26 & 0.632 & 0.362 & 17.59 & 0.516 & 0.325 & 21.12 & 0.837 & 0.372 & 18.39 & 0.614 & 0.465 \\
        & M.-Nano~\citep{brateanu2026multinex} & 19.73 & 0.604 & 0.357 & 16.87 & 0.506 & 0.354 & 19.28 & 0.807 & 0.419 & 16.93 & 0.572 & 0.443 \\
        \midrule
        \multirow{10}{*}{\rotatebox{90}{\textbf{UL}}}
        & Zero-DCE~\citep{guo2020zero} & 16.40 & 0.475 & \bestresult{0.368} & 15.03 & 0.418 & 0.244 & 17.07 & 0.639 & 0.514 & 13.88 & 0.542 & 0.436 \\
        & RUAS~\citep{liu2021retinex} & 15.71 & 0.501 & 0.493 & 12.14 & 0.438 & 0.413 & 11.69 & 0.632 & 0.519 & 7.80 & 0.432 & 0.726 \\
        & SCI~\citep{ma2022toward} & 15.70 & 0.439 & 0.379 & 14.55 & 0.410 & 0.241 & 15.47 & 0.598 & 0.531 & 9.68 & 0.479 & 0.580 \\
        & PairLIE~\citep{fu2023learning} & 18.98 & 0.562 & 0.377 & 15.52 & 0.435 & 0.257 & 17.53 & 0.663 & 0.456 & 12.47 & 0.512 & 0.497 \\
        & CLIP-LIT~\citep{liang2023iterative} & 13.56 & 0.424 & 0.405 & 13.37 & 0.382 & 0.276 & 15.27 & 0.590 & 0.551 & 13.45 & 0.530 & 0.416 \\
        & CoLIE~\citep{chobola2024fast} & 14.76 & 0.420 & 0.401 & 12.87 & 0.389 & 0.277 & 17.89 & 0.625 & - & 14.06 & 0.543 & \secondresult{0.409} \\
        & CLODE~\citep{jung2025continuous} & 18.44 & 0.564 & 0.482 & 15.54 & 0.487 & 0.388 & 16.17 & \secondresult{0.777} & 0.454 & 13.49 & 0.514 & 0.567 \\
        & LASQ~\citep{kong2026luminance} & 19.67 & \bestresult{0.599} & 0.389 & 16.12 & \bestresult{0.491} & 0.257 & 20.13 & 0.749 & 0.399 & 14.01 & 0.540 & 0.450 \\
        \cmidrule(l){2-14}
        & \textbf{RISE} & \bestresult{20.45} & \secondresult{0.590} & 0.398 & \bestresult{16.71} & \bestresult{0.491} & \bestresult{0.236} & \bestresult{22.22} & 0.771 & \secondresult{0.384} & \secondresult{14.80} & \secondresult{0.554} & 0.466 \\
        & \textbf{RISE}$^\dagger$ & \secondresult{20.24} & 0.580 & \secondresult{0.374} & \bestresult{16.71} & \bestresult{0.491} & \bestresult{0.236} & \secondresult{22.03} & \bestresult{0.788} & \bestresult{0.373} & \bestresult{16.73} & \bestresult{0.570} & \bestresult{0.407} \\
        \bottomrule
\end{tabular*}
\vspace{-3mm}
\end{table*}

\begin{table*}[t]
    \centering
    \caption{No-reference quantitative comparison on unpaired datasets. B. denotes BRISQUE.}
    \label{tab:no-reference-comparison}
    \fontsize{8}{9.5}\selectfont
    \setlength{\tabcolsep}{0.4pt}
    \renewcommand{\arraystretch}{1.0}
    \begin{tabular*}{\textwidth}{@{\extracolsep{\fill}}c l ccc ccc ccc ccc ccc@{}}
        \toprule
        \multirow{2}{*}{\textbf{Type}}
        & \multirow{2}{*}{\textbf{Method}}
        & \multicolumn{3}{c}{\textbf{DICM}}
        & \multicolumn{3}{c}{\textbf{LIME}}
        & \multicolumn{3}{c}{\textbf{MEF}}
        & \multicolumn{3}{c}{\textbf{NPE}}
        & \multicolumn{3}{c}{\textbf{VV}} \\
        \cmidrule(lr){3-5}\cmidrule(lr){6-8}\cmidrule(lr){9-11}
        \cmidrule(lr){12-14}\cmidrule(lr){15-17}
	        & & IQA$\uparrow$ & IAA$\uparrow$ & B.$\downarrow$
	        & IQA$\uparrow$ & IAA$\uparrow$ & B.$\downarrow$
	        & IQA$\uparrow$ & IAA$\uparrow$ & B.$\downarrow$
	        & IQA$\uparrow$ & IAA$\uparrow$ & B.$\downarrow$
	        & IQA$\uparrow$ & IAA$\uparrow$ & B.$\downarrow$ \\
        \midrule
        \multirow{2}{*}{\rotatebox{90}{\textbf{SL}}}
        & MIRNet~\citep{zamir2020learning}
	        & 2.69 & 1.90 & 42.33 & 2.40 & 1.78 & 35.68 & 2.17 & 1.57 & 42.34
	        & 2.21 & 1.71 & 31.01 & 2.58 & 1.67 & 31.43 \\
        & CIDNet~\citep{yan2025hvi}
	        & 2.87 & 2.05 & 34.00 & 2.76 & 2.12 & \bestresult{17.25} & 2.75 & 2.20 & \bestresult{20.20}
	        & 2.88 & 1.87 & 28.18 & 2.84 & 1.85 & \bestresult{13.63} \\
        \midrule
        \multirow{5}{*}{\rotatebox{90}{\textbf{UL}}}
        & RUAS~\citep{liu2021retinex}
	        & 2.04 & 1.77 & 47.50 & 2.11 & 1.85 & 29.26 & 2.69 & 2.11 & 32.49
	        & 1.66 & 1.42 & 49.94 & 2.27 & 1.66 & 40.48 \\
        & Li et al.~\citep{li2025interpretable}
	        & 2.73 & 1.89 & 31.94 & 2.38 & 1.82 & 18.48 & 2.48 & 1.85 & \secondresult{23.68}
	        & 2.87 & 1.92 & 28.74 & 2.64 & 1.75 & 20.78 \\
        & CLODE~\citep{jung2025continuous}
	        & 2.48 & 1.71 & 36.80 & 2.43 & 1.79 & 29.93 & 2.30 & 1.72 & 33.84
	        & 2.42 & 1.58 & 32.70 & 2.56 & 1.60 & 28.47 \\
        & LASQ~\citep{kong2026luminance}
	        & \secondresult{3.43} & \secondresult{2.26} & \secondresult{30.20}
	        & \secondresult{2.98} & \secondresult{2.18} & \secondresult{17.37}
	        & \secondresult{3.04} & \secondresult{2.23} & 25.24
	        & \secondresult{3.43} & \secondresult{2.20} & \secondresult{23.33}
	        & \secondresult{3.37} & \secondresult{2.02} & 17.24 \\
        \cmidrule(l){2-17}
        & \textbf{RISE}
	        & \bestresult{3.73} & \bestresult{2.47} & \bestresult{30.00}
	        & \bestresult{3.42} & \bestresult{2.54} & 21.54
	        & \bestresult{3.45} & \bestresult{2.44} & 24.95
	        & \bestresult{3.57} & \bestresult{2.45} & \bestresult{21.48}
	        & \bestresult{3.47} & \bestresult{2.06} & \secondresult{14.33} \\
        \bottomrule
    \end{tabular*}
\end{table*}

\begin{figure*}[t]
    \centering
    \includegraphics[width=\textwidth]{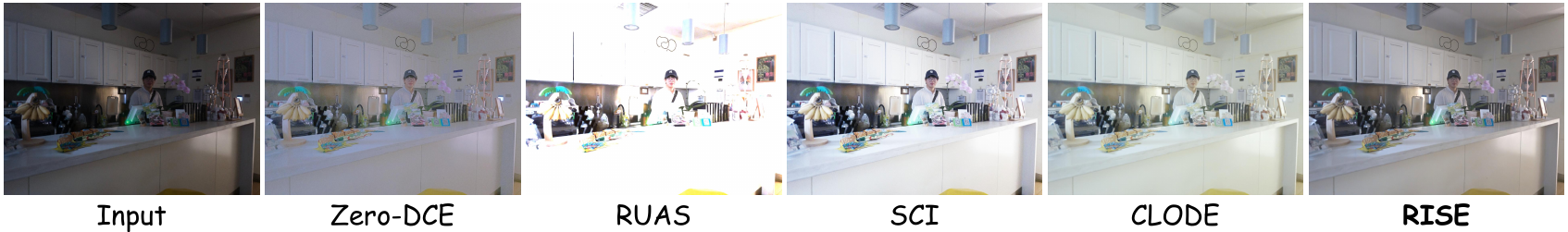}
    \caption{Generalization to real-world scenes captured with an iPhone 16 Pro. Zoom in for
    the best view.}
    \label{fig:reality}
\end{figure*}

\begin{table*}[t]
    \centering
    \begingroup
    \fontsize{8}{9.5}\selectfont
    \setlength{\tabcolsep}{0.6pt}
    \renewcommand{\arraystretch}{0.7}
    \caption{Ablation studies of relative structure, KIC selection, DMER, decoupling, and losses.}
    \label{tab:ablation}
    \label{tab:ablation-study}
    \label{tab:ablation-relation}
    \begin{tabular*}{\textwidth}{@{\extracolsep{\fill}}lccc@{\hspace{5pt}}lccc@{\hspace{5pt}}lccc@{}}
        \toprule
        \multicolumn{4}{c}{\textbf{(a) Relative Structure}} &
        \multicolumn{4}{c}{\textbf{(b) KICs and DMER}} &
        \multicolumn{4}{c}{\textbf{(c) Decoupling and Loss}} \\
        \cmidrule(lr){1-4}\cmidrule(lr){5-8}\cmidrule(lr){9-12}
        \textbf{Removed term} & PSNR$\uparrow$ & SSIM$\uparrow$ & LPIPS$\downarrow$ &
        \textbf{Setting} & PSNR$\uparrow$ & SSIM$\uparrow$ & LPIPS$\downarrow$ &
        \textbf{Setting} & PSNR$\uparrow$ & SSIM$\uparrow$ & LPIPS$\downarrow$ \\
        \midrule
        $v_n\rho_n(p)Z_{k_n}$ & 15.74 & 0.480 & 0.257 &
        Random KICs & 15.84 & 0.480 & 0.248 &
        w/o $\mathbf{S}$ & 15.16 & 0.476 & 0.256 \\
        $u_n\rho_n(p)\delta_n(p)$ & 16.09 & 0.485 & 0.248 &
        Darkest KICs & 14.49 & 0.482 & 0.278 &
        w/o $\eta$ & 15.47 & 0.489 & 0.264 \\
        $w_n\delta_n(p)$ & 15.46 & 0.477 & 0.242 &
        $\mathbf{T}_\mathrm{abs}$ & 14.92 & 0.471 & 0.269 &
        w/o $\mathcal{L}_{\mathrm{sc}}$ & 16.04 & 0.485 & 0.249 \\
        $\boldsymbol{\beta}^{\top}\mathbf{c}_{\mathrm{scene}}+b$ & 15.96 & 0.482 & 0.244 &
        $\mathbf{T}_\mathrm{rel}$ & 15.36 & 0.482 & 0.251 &
        w/o $\mathcal{L}_{\mathrm{idt}}$ & 16.40 & 0.484 & 0.247 \\
        \midrule
        \rowcolor{rankfirst}[6pt][6pt]
        \textbf{None (Full)} & \textbf{16.71} & \textbf{0.491} & \textbf{0.236} &
        \textbf{Brightest+DMER} & \textbf{16.71} & \textbf{0.491} & \textbf{0.236} &
        \textbf{Full} & \textbf{16.71} & \textbf{0.491} & \textbf{0.236} \\
        \bottomrule
\end{tabular*}
    \endgroup
\end{table*}

\begin{table*}[!htbp]
    \centering
    \begingroup
    \fontsize{8}{9.5}\selectfont
    \setlength{\tabcolsep}{0.6pt}
    \renewcommand{\arraystretch}{0.95}
    \caption{Ablation studies on the number and size of KICs and sensitivity analysis of the blending weight $\alpha$ in DMER.}
    \label{tab:kic-number-size}
    \label{tab:dmer-alpha-sensitivity}
    \begin{tabular*}{\textwidth}{@{\extracolsep{\fill}}ccc@{\hspace{5pt}}ccc@{\hspace{5pt}}cccccc@{}}
        \toprule
        \multicolumn{3}{c}{\textbf{(a) Number of KICs}} &
        \multicolumn{3}{c}{\textbf{(b) KIC Size}} &
        \multicolumn{6}{c}{\textbf{(c) DMER Blending Weight}} \\
        \cmidrule(lr){1-3}\cmidrule(lr){4-6}\cmidrule(lr){7-12}
        $N$ & PSNR$\uparrow$ & SSIM$\uparrow$ &
        Size & PSNR$\uparrow$ & SSIM$\uparrow$ &
        $\boldsymbol{\alpha}$ & PSNR$\uparrow$ & SSIM$\uparrow$ &
        $\boldsymbol{\alpha}$ & PSNR$\uparrow$ & SSIM$\uparrow$ \\
        \midrule
        1 & 16.65 & 0.492 & $50\times25$ & 16.66 & 0.492 & 0.3 & 16.50 & 0.490 & 0.6 & 16.41 & 0.490 \\
        5 & 16.81 & 0.492 & $150\times100$ & 16.59 & 0.481 & 0.4 & \bestresult{16.71} & \bestresult{0.491} & 0.7 & 16.10 & 0.488 \\
        3 & 16.71 & 0.491 & $75\times50$ & 16.71 & 0.491 & 0.5 & 16.48 & \bestresult{0.491} &     &       &       \\
        \bottomrule
    \end{tabular*}
    \endgroup
\end{table*}

To translate $\mathbf{L}$ into an exposure adjustment, we adopt the Exposure-Value (EV) scale used in photography, where \emph{one stop corresponds to a twofold change in exposure}. Following the conventional normal-key setting in photographic tone reproduction~\citep{reinhard2002photographic}, we use $m=0.18$ as the 0-stop reference. The observed luminance $\mathbf{Y}$ and the structure-aware light field $\mathbf{L}$ provide complementary exposure corrections, which are naturally additive in EV stops:
\begin{equation}
    e_{\mathrm{rel}}
    =
    \log_2\!\frac{\operatorname{mean}(\mathbf{Y})}{m}
    +
    \log_2\!\frac{\operatorname{mean}(\mathbf{L})}{m}.
    \label{eq:dmerev}
\end{equation}

Because the raw EV compensation can become excessive in severely underexposed scenes, a range-control function $\Phi(\cdot)$ constrains it before conversion to a linear gain, yielding the relative reference:
\begin{equation}
    \mathbf{T}_{\mathrm{rel}}
    = 2^{\Phi(e_{\mathrm{rel}})}\,\mathbf{I}_{\mathrm{low}}.
    \label{eq:dmerrel}
\end{equation}

We blend the two references and apply Gaussian smoothing to
obtain the final exposure reference:
\begin{equation}
    \mathbf{T} = \mathcal{G}_{\sigma}\!\left(
    (1-\alpha)\,\mathbf{T}_{\mathrm{abs}}
    + \alpha\,\mathbf{T}_{\mathrm{rel}}
    \right),
    \label{eq:dmerfuse}
\end{equation}
where $\alpha=0.4$ balances the two references, and $\mathcal{G}_{\sigma}$ suppresses high-frequency fluctuations while preserving low-frequency brightness structure. Since both references are input-dependent, $\mathbf{T}$ provides scene-conditioned supervision rather than a fixed target, encouraging RISE to learn exposure mappings that generalize across illumination distributions.

\subsection{Learning Objective}

\noindent\textbf{Exposure control loss.}
Let $\mathbf{G}_{\mathbf{T}}$ and $\boldsymbol{\ell}_{\mathbf{T}}$ be the patch
log-gain and KIC luminance derived from $\mathbf{T}$. The exposure loss aligns
the output with $\mathbf{T}$ at three levels,
\begin{equation}
    \mathcal{L}_{\mathrm{exp}} =
    \|\mathbf{Y}_{\hat{\mathbf{I}}}-\mathbf{Y}_{\mathbf{T}}\|_1
    + \|\hat{\mathbf{G}}-\mathbf{G}_{\mathbf{T}}\|_1
    + \|\hat{\boldsymbol{\ell}}-\boldsymbol{\ell}_{\mathbf{T}}\|_1.
    \label{eq:lossexp}
\end{equation}

\noindent\textbf{Regularizers.}
Scale equivariance preserves relative structure under global intensity changes. Given $(\Theta_s,\eta_s)$ from scaled input $s\mathbf{I}_{\mathrm{low}}$, structural parameters remain stable while $\eta$ absorbs the change,
\begin{equation}
    \mathcal{L}_{\mathrm{sc}} =
    \|\Theta_s-\Theta\|_1
    + \|\eta_s-\eta+\log s\|_1.
    \label{eq:losssc}
\end{equation}
An idempotence regularizer $\mathcal{L}_{\mathrm{idt}}
= \|\mathbf{G}_\mathrm{e}\|_1 + \|\eta_\mathrm{e}\|_1
+ \|\boldsymbol{\ell}_\mathrm{e}-\hat{\boldsymbol{\ell}}\|_1$ discourages further
enhancement on already enhanced outputs $\hat{\mathbf{I}}$, yielding the re-enhanced result $\mathbf{I}_\mathrm{e}$, while an edge-aware smoothness loss
$\mathcal{L}_{\mathrm{smooth}}$ suppresses gain variations unsupported by
luminance edges. The overall objective is:
\begin{equation}
    \mathcal{L} = \lambda_{\mathrm{exp}}\mathcal{L}_{\mathrm{exp}}
    + \lambda_{\mathrm{sc}}\mathcal{L}_{\mathrm{sc}}
    + \lambda_{\mathrm{idt}}\mathcal{L}_{\mathrm{idt}}
    + \lambda_{\mathrm{smooth}}\mathcal{L}_{\mathrm{smooth}}.
    \label{eq:losstotal}
\end{equation}
We empirically set the loss weights $\lambda_{\mathrm{exp}},\lambda_{\mathrm{sc}},
\lambda_{\mathrm{idt}},\lambda_{\mathrm{smooth}}=[1.0,0.1,0.05,0.01]$.

\section{Experimental Validation}

\subsection{Datasets and Experimental Details}

\noindent\textbf{Datasets.} We evaluate RISE on seven paired benchmarks and five unpaired datasets. LOL-v1 contains 485 training and 15 testing pairs~\citep{wei2018retinexnet}. LOLv2-Real includes 689/100 training/testing pairs, while LOLv2-Synthetic includes 900/100~\citep{yang2020fidelity,yang2021sparse}. The Huawei and Nikon subsets of LSRW contain 2{,}450/30 and 3{,}150/20 training/testing pairs, respectively~\citep{hai2023r2rnet}. UHD-LL provides 2{,}000 training and 150 testing pairs at 4K resolution~\citep{wang2023uhd}. SICE~\citep{cai2018learning} contains 4{,}413 images from 589 high-resolution multi-exposure sequences; we use all 360 Part~1 sequences for training and all 229 Part~2 sequences for testing. For no-reference evaluation, we use DICM~\citep{lee2013contrast}, LIME~\citep{guo2016lime}, MEF~\citep{ma2015perceptual}, NPE~\citep{wang2013naturalness}, and VV~\citep{vonikakis2018evaluation}.

\noindent\textbf{Metrics.} For reference-based evaluation, we report PSNR,
SSIM~\citep{wang2004image}, and LPIPS with AlexNet~\citep{zhang2018perceptual}.
For no-reference evaluation, we use the IQA and IAA scores from Q-Align~\citep{wu2024qalign}, together with BRISQUE~\citep{mittal2012making}.

\noindent\textbf{Implementation Details.} RISE is trained for 100 epochs on a single NVIDIA V100 GPU with a batch size of 8. We use Adam~\citep{kingma2015adam} with an initial learning rate of $2\times10^{-4}$. The patch size is $75\times50$, and we set $N=3$. RISE is trained only on LSRW-Nikon, whereas RISE$^\dagger$ is trained separately on each benchmark. We use official LSRW-Nikon checkpoints when available, otherwise, we use official weights trained on their optimal training sets.

\subsection{Comparisons with State-of-the-Art Methods}

\noindent\textbf{Quantitative Evaluation.}
We compare RISE with representative supervised and unsupervised methods in Tables~\ref{tab:quantitative-comparison-LOL} and~\ref{tab:quantitative-comparison-LSRW}. RISE achieves the highest PSNR among unsupervised methods on all datasets. Its consistent performance without dataset-specific retraining indicates transferable exposure regularity rather than overfitting to particular brightness distributions. We further evaluate generalization on five unpaired benchmarks in Table~\ref{tab:no-reference-comparison}, using the same checkpoint trained on LSRW-Nikon. RISE achieves the highest IQA and IAA scores across all datasets by clear margins, demonstrating perceptually natural enhancement on unseen real-world scenes. Moreover, RISE requires only 0.35M parameters and 1.43 GFLOPs, highlighting its computational efficiency.

\noindent\textbf{Qualitative Evaluation.}
The visual results in Fig.~\ref{fig:lolresults} indicate that RISE produces more natural exposure correction with well-preserved color fidelity. Furthermore, RISE achieves adaptive enhancement effects across different scenes, while other methods exhibit evident fixed enhancement styles. For more visual comparisons based on the benchmarks, please refer to supplementary material.

\noindent\textbf{Real-world Generalization.}
To evaluate generalization beyond benchmarks, we apply the LSRW-Nikon checkpoint to low-light photos captured with an iPhone 16 Pro, as shown in Figure~\ref{fig:reality}. Without additional tuning, RISE produces natural exposure with a broader dynamic range, recovering shadow and highlight details while preserving faithful colors. More results are provided in supplementary material.

\subsection{Ablation Study}

\noindent\textbf{Relative Structure.}
Table~\ref{tab:ablation-relation}(a) reveals the interplay among the relational
terms. Ablations show that direct discrepancies,
distance-aware relations, and scene context jointly form the coherent
scene-wide relative illumination structure.

\noindent\textbf{KICs and DMER.}
Table~\ref{tab:ablation-study}(b) evaluates KIC selection and DMER, while Table~\ref{tab:kic-number-size}(a--b) examines the number and size of KICs. Random or darkest KICs degrade performance, especially the latter, supporting bright regions as reliable illumination cues. In contrast, varying the number or size of KICs yields comparable performance, indicating that RISE is largely insensitive to these configurations. Using either meter alone also degrades performance, confirming their complementarity in global calibration and spatial illumination adaptation. Table~\ref{tab:dmer-alpha-sensitivity}(c) further shows that DMER is robust to variations in $\alpha$: $\alpha=0.4$ achieves the best PSNR, while $\alpha\in[0.3,0.5]$ yields comparable performance. The degradation at larger $\alpha$ indicates that retaining a slightly larger contribution from the absolute meter better balances global exposure calibration and scene-adaptive spatial adjustment.

\noindent\textbf{Decoupling and Loss.}
Table~\ref{tab:ablation-relation}(c) ablates both our formulation of the
enhancement problem and the regularization terms in the objective. Relative
structure alone cannot provide sufficient exposure adjustment, whereas absolute
exposure alone cannot model spatial variations in illumination. The performance
drop caused by removing $\mathcal{L}_{\mathrm{sc}}$ further supports our premise
that scene illumination structure should remain stable under global intensity
changes, while such changes should be absorbed by the absolute exposure term
rather than structural parameters.

\FloatBarrier

\section{Conclusion and Limitation}

This paper proposes RISE, formulating low-light enhancement as relative illumination structure estimation and absolute exposure control. RISE derives structure from reliable KIC-based spatial relations, while DMER provides scene-conditioned unsupervised guidance. Extensive experiments consistently demonstrate leading unsupervised performance and strong generalization across lighting conditions and unseen real-world scenes. However, when signals fall below the sensor noise floor, missing content cannot be recovered through exposure correction alone, as shown by failure cases in the supplementary material. Incorporating generative restoration remains a promising future direction.

\bibliography{rise_references}
\bibliographystyle{rise_conference}

\clearpage
\appendix
\renewcommand{\thesection}{A.\arabic{section}}
\setcounter{section}{0}


\section{Relative Illumination beyond Brightness Ordering}
\label{app:brightness-ordering}

We clarify why RISE models continuous, spatially conditioned illumination
relations instead of directly adopting the regional brightness ordering observed
in a low-light input. The analysis first examines whether the input ordering
reliably describes the desired normal-light appearance, and then highlights the
additional information captured by our relative illumination formulation.

The relative brightness ordering in a low-light observation is not necessarily
consistent with that in its normally exposed counterpart. Regional luminance is
not a direct measurement of illumination, but is jointly determined by
illumination, surface reflectance, sensor noise, quantization, and the camera
response. In particular, weak measurements in severely underexposed regions
can compress or even reverse pairwise brightness relations. Directly treating
the input ordering as the target illumination structure may therefore impose
incorrect ordinal constraints on the enhanced result. We analyze this
phenomenon using paired low-light and ground-truth images only for evaluation;
the ground truth is never used to train RISE.

\begin{figure}[!htbp]
    \centering
    \includegraphics[width=\textwidth]{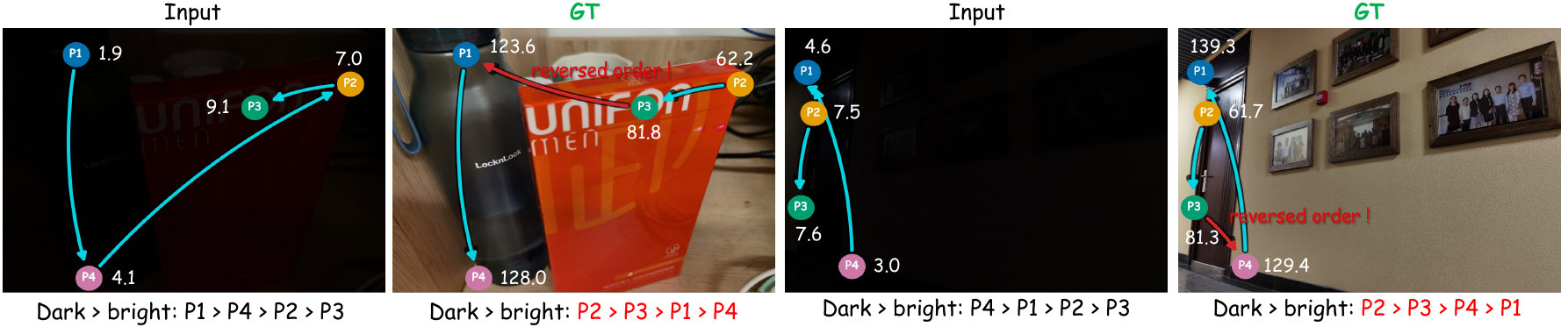}
    \caption{Inconsistency of relative brightness ordering.
    Representative examples where the regional brightness ordering in
    low-light inputs differs from that in the corresponding ground-truth
    images, indicating that input brightness order is not a reliable target
    illumination constraint.}
    \label{fig:appendix-bright-order}
\end{figure}

\begin{table*}[!htbp]
    \centering
    \begingroup
    \setlength{\tabcolsep}{1.5pt}
    \renewcommand{\arraystretch}{1.05}
    \caption{Quantitative comparison of luminance distributions. Values represent the Wasserstein Distance ($\downarrow$) to the corresponding ground truth. The best and second-best results are highlighted in \colorbox{rankfirst}{\textbf{orange}} and \colorbox{ranksecond}{yellow}, respectively, with the best results also shown in boldface.}
    \label{tab:appendix-luminance-distribution}
    \small
    \begin{tabular*}{\textwidth}{@{\extracolsep{\fill}}lcccccc@{}}
        \toprule
        \textbf{Method} & \textbf{LOL-v1} & \textbf{LOLv2-Real} &
        \textbf{LOLv2-Syn} & \textbf{LSRW-Huawei} &
        \textbf{LSRW-Nikon} & \textbf{UHD-LL} \\
        \midrule
        Zero-DCE~\citep{guo2020zero} & 0.1684 & 0.1070 & 0.0907 & 0.1223 & 0.0923 & 0.0744 \\
        RUAS~\citep{liu2021retinex} & 0.0792 & 0.0890 & 0.1022 & 0.0928 & 0.1300 & 0.2075 \\
        SCI~\citep{ma2022toward} & 0.1816 & 0.1179 & 0.0618 & 0.1402 & \secondresult{0.0572} & 0.1071 \\
        CLIP-LIT~\citep{liang2023iterative} & 0.2374 & 0.1587 & 0.1325 & 0.1797 & 0.1589 & 0.1520 \\
        CoLIE~\citep{chobola2024fast} & 0.2087 & 0.1643 & 0.1961 & 0.1623 & 0.1922 & 0.1414 \\
        CLODE~\citep{jung2025continuous} & \bestresult{0.0266} & 0.0503 & 0.0636 & 0.0475 & 0.0829 & 0.1186 \\
        \midrule
        \textbf{RISE} & 0.0769 & \bestresult{0.0213} & \secondresult{0.0348} & \bestresult{0.0305} & \bestresult{0.0281} & \bestresult{0.0271} \\
        \textbf{RISE}$^\dagger$ & \secondresult{0.0609} & \secondresult{0.0422} & \bestresult{0.0345} & \secondresult{0.0351} & \bestresult{0.0281} & \secondresult{0.0525} \\
        \bottomrule
\end{tabular*}
    \endgroup
\end{table*}

These inconsistencies indicate that input brightness relations should serve as
evidence for illumination estimation rather than constraints to be directly
preserved. A rank-based representation retains only whether one region is
brighter than another. In contrast, the discrepancy used by RISE satisfies
$\delta_n(p)=Z_{k_n}-Z_p=\log(P_{k_n}/P_p)$, preserving the magnitude of the
relative brightness ratio while remaining invariant to global intensity
scaling. RISE further combines this continuous relation with spatial
attenuation $\rho_n(p)$, multiple KICs, predicted KIC luminance, and scene
context to construct a scene-conditioned gain field. Importantly, KICs are
selected as high-SNR anchors rather than fixed ordering targets. Their
relations guide the estimated correction without enforcing the input
brightness ordering on the enhanced output. Consequently, RISE can distinguish
illumination patterns that share the same ordinal ranking but exhibit different
attenuation strengths, while correcting unreliable relations inherited from
low-SNR observations.

\section{Details of the Transition from Log-Gain to Image-Space Enhancement}
\label{app:log-gain-transition}

We further clarify the transition from Eq.~(4) to Eq.~(5) in the main paper.
RISE does not predict an additive offset to the image intensity. Instead, it
predicts the logarithm of a multiplicative exposure gain. At the patch level,
Eq.~(4) decomposes this log-gain into relative illumination structure and
absolute exposure:
\begin{equation*}
    \hat{G}(p)=S(p)+\eta.
\end{equation*}
Before it is applied to the input image, the patch-level field is upsampled to
the pixel grid and mapped back to the linear-gain domain through the
exponential function. For a pixel $x$, the resulting gain is
\begin{align*}
    g(x)
    = \exp\!\left(\operatorname{Up}(\hat{\mathbf{G}})(x)\right) 
    = \exp\!\left(\operatorname{Up}(\mathbf{S})(x)+\eta\right) 
    = \exp\!\left(\operatorname{Up}(\mathbf{S})(x)\right)\exp(\eta).
\end{align*}
The enhanced image in Eq.~(5) is therefore obtained by multiplying this gain
with the low-light input:
\begin{equation*}
    \hat{\mathbf{I}}(x)=\mathbf{I}_{\mathrm{low}}(x)\odot g(x).
\end{equation*}
This derivation explains why the two components are additive in log-gain space
but multiplicative in the linear-gain domain: the relative term determines the
spatially varying correction, whereas $\eta$ scales the exposure of the entire
image. The parameterization also gives a direct interpretation of the predicted
values. Specifically, $\hat{G}=0$ yields $g=1$ and leaves the intensity
unchanged; $\hat{G}=\log 2$ yields $g=2$ and doubles it; and
$\hat{G}=\log 0.5$ yields $g=0.5$ and halves it. More generally, positive
log-gain values enhance exposure, while negative values suppress it.



\section{Quantitative Luminance Distribution Comparison}
\label{app:luminance-distribution}

Beyond standard image restoration metrics, we further examine the luminance
distributions produced by different unsupervised LLIE methods. We compute the
Wasserstein Distance between the luminance distribution of each enhanced result
and that of the corresponding ground truth. A smaller distance indicates that
the enhanced image better matches the target exposure distribution. As shown in
Table~\ref{tab:appendix-luminance-distribution}, RISE obtains the lowest or
near-lowest distance on most benchmarks, suggesting that the proposed
illumination modeling yields more faithful luminance distributions across
diverse datasets.

\section{More Qualitative Results}
\label{app:qualitative}

Additional qualitative results on LOL, LSRW, UHD-LL, and SICE complement the
main comparisons. These examples cover diverse illumination distributions and
emphasize both the recovery of poorly illuminated content and the preservation
of details in bright regions.

\begin{figure}[!htbp]
    \centering
    \includegraphics[width=\textwidth]{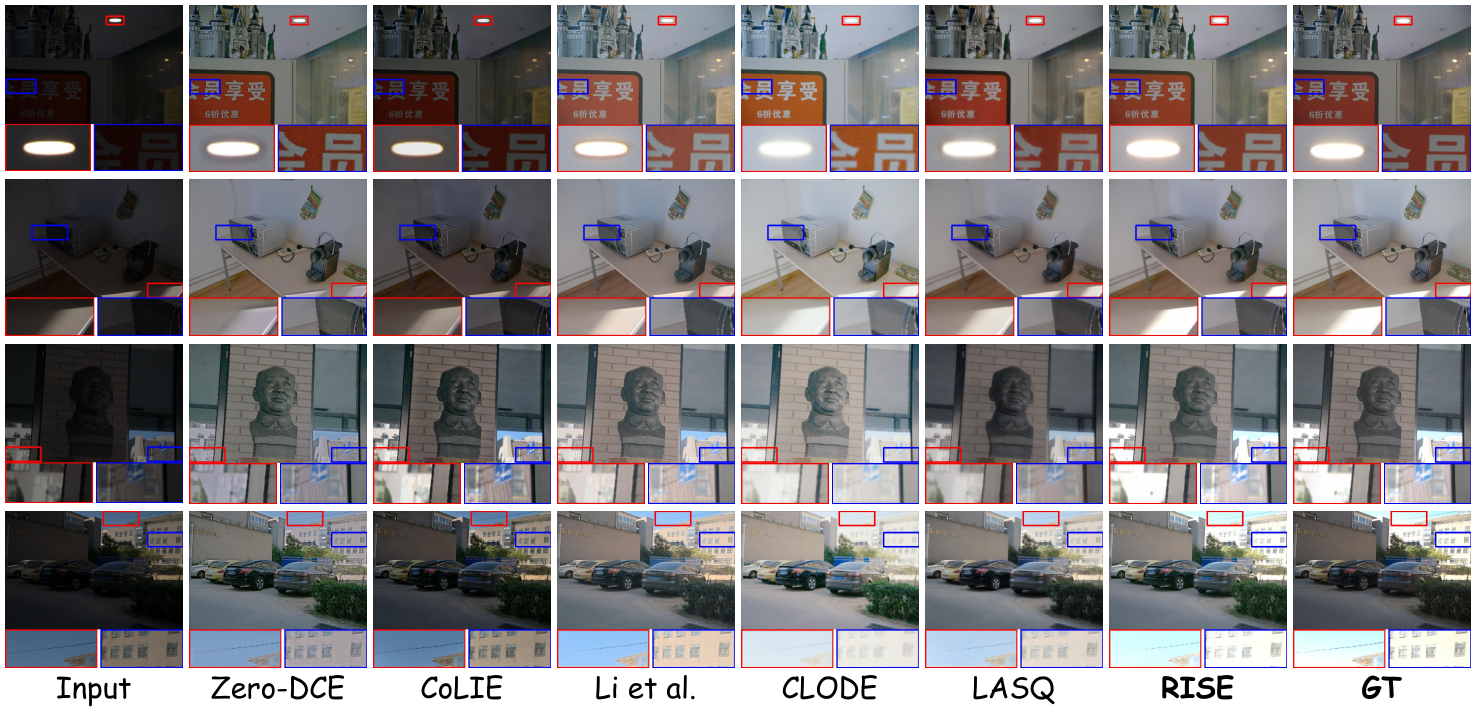}
    \caption{Additional visual comparisons on the LOL benchmarks.}
    \label{fig:appendix-lol}
\end{figure}

\begin{figure}[!htbp]
    \centering
    \includegraphics[width=\textwidth]{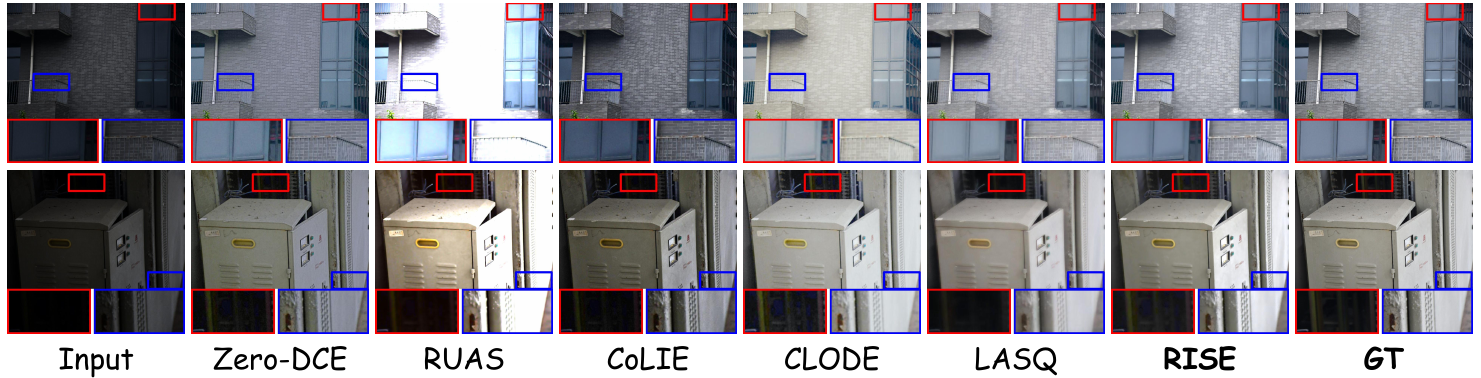}
    \caption{Additional visual comparisons on LSRW.}
    \label{fig:appendix-lsrw}
\end{figure}

\begin{figure}[!htbp]
    \centering
    \includegraphics[width=\textwidth]{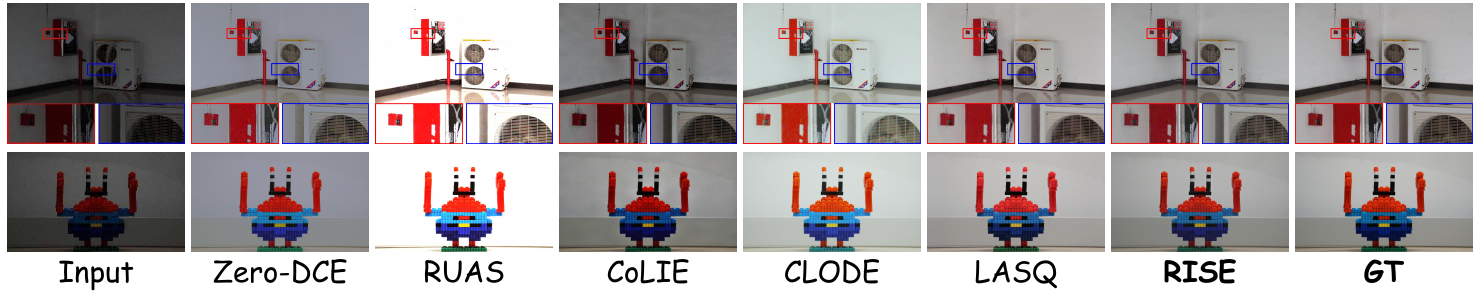}
    \caption{Additional visual comparisons on UHD-LL.}
    \label{fig:appendix-uhdll}
\end{figure}

\begin{figure}[!htbp]
    \centering
    \includegraphics[width=\textwidth]{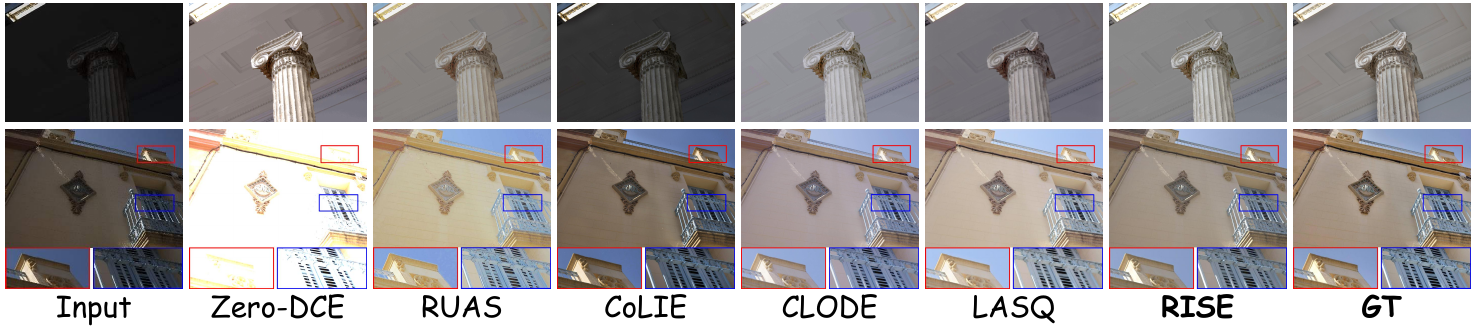}
    \caption{Additional visual comparisons on SICE.}
    \label{fig:appendix-sice}
\end{figure}

\begin{figure}[!htbp]
    \centering
    \includegraphics[width=\textwidth]{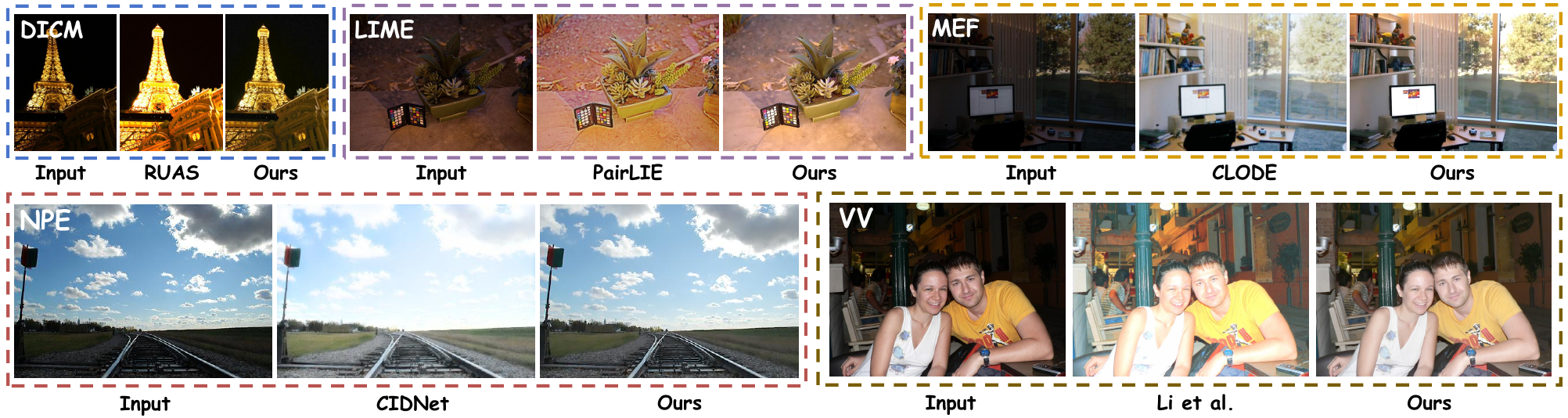}
    \caption{Visual comparisons on No-Reference Datasets.}
    \label{fig:appendix-unpaired}
\end{figure}

\section{More Real-world Generalization Results}
\label{app:real-world}

For out-of-benchmark generalization, we apply the same LSRW-Nikon checkpoint to
low-light photographs captured with an iPhone 16 Pro, without additional
training or scene-specific tuning.

\begin{figure}[!htbp]
    \centering
    \includegraphics[width=\textwidth]{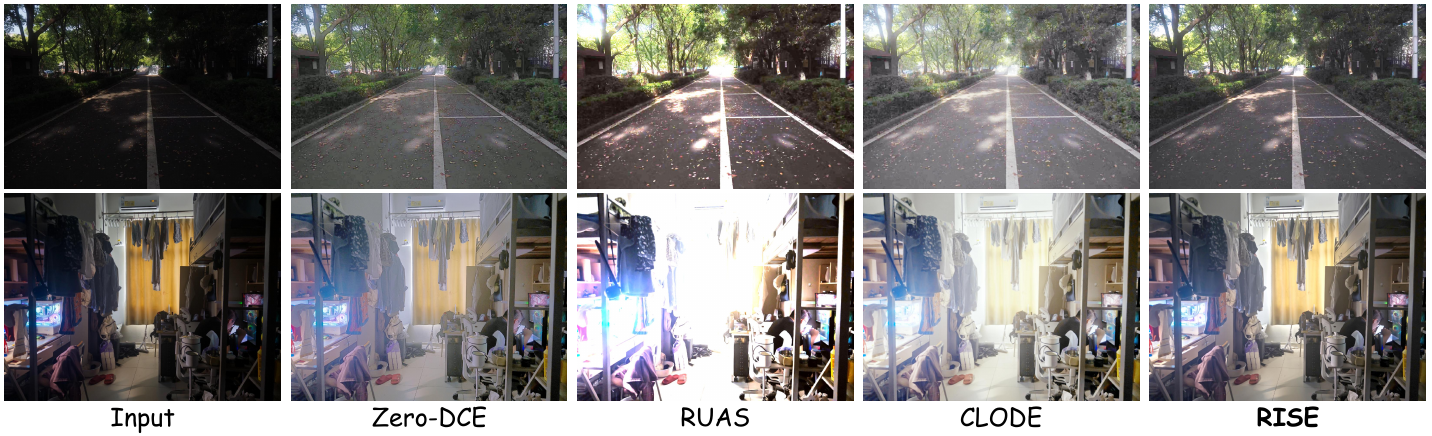}
    \caption{Additional comparisons on real-world low-light scenes.
    RISE provides balanced exposure in both outdoor and indoor scenes while
    avoiding severe highlight saturation. All results are produced by the
    LSRW-Nikon checkpoint without additional tuning.}
    \label{fig:appendix-reality}
\end{figure}

\section{Failure Cases}
\label{app:failure}

RISE estimates illumination from the visual evidence retained in the input and
cannot reconstruct scene content that falls below the sensor noise floor.
Representative examples illustrate this limitation in extremely dark regions.

\begin{figure}[!htbp]
    \centering
    \includegraphics[width=\textwidth]{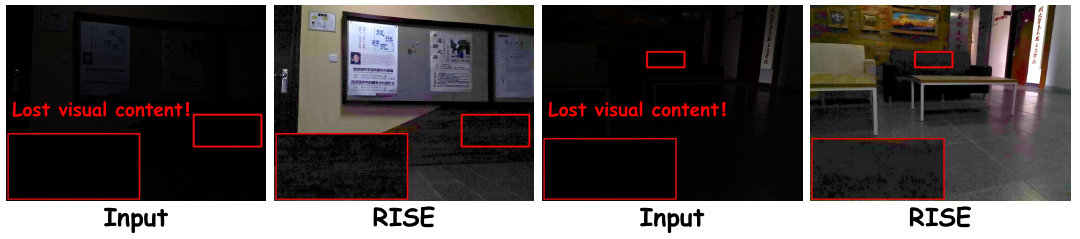}
    \caption{Failure cases under unrecoverable signal loss.
    RISE improves overall visibility but cannot recover structures absent from
    the recorded input and may retain residual noise in these regions. Red
    boxes and enlarged crops indicate the affected areas.}
    \label{fig:appendix-failure}
\end{figure}

\end{document}